\documentclass[journal,twoside,web]{ieeecolor}
\usepackage{lcsys}

\usepackage{graphics,graphicx} 
\usepackage{amsmath,float} 
\usepackage{amssymb}  
\usepackage[export]{adjustbox}
\usepackage{float}
\usepackage{array}

\title{\LARGE  Efficiency Optimization of a Two-link Planar Robotic Arm}

\author{Meysam Fathizadeh$^{1}$ and Hanz Richter$^{2}$ %
\thanks{*This work was supported by NSF grant \#2221726.}
\thanks{$^{1}$sResearch assistant, Mechanical Engineering Department, Cleveland State University {\tt\small m.fathizadeh41@vikes.csuohio.edu}.}
\thanks{$^{2}$Professor, Mechanical Engineering Department, Cleveland State University {\tt\small h.richter@csuohio.edu}.}
}

\begin{document}

\maketitle
\thispagestyle{empty}
\pagestyle{empty}

\begin{abstract}
Energy consumption optimization of a two-link planar robotic arm is considered with the system's efficiency being the target for optimization. A new formulation of thermodynamic principles within the framework of dynamical systems is used. This approach is applied by considering cyclic motions for the robotic arm and analyzing the cyclic averaged energies while the robotic arm is tasked with going from point A to point B in the task space while resisting an external force. The energy transfer rate between the links is classified into positive and negative and the results combined with the averaged energy quantities, are used to address the optimization problem while adhering to the constraints imposed by the second law of thermodynamics in its new formulation.

\end{abstract}
\section{Introduction and Problem Setting}

Understanding and analyzing hybrid multi-physics systems through the lens of thermodynamic principles provides a comprehensive approach to studying their behavior \cite{arent2021multi}.\\
It is shown by H. Richter~\cite{IEEELCS22} that Hamiltonian systems that work in cycles show behaviors in agreement with the second law of thermodynamics if the averaged energies were considered. The concept of Energy Cyclo-Directionality (ECD) introduced in ~\cite{IEEELCS22} opens the way to benefit from all the principles available in the thermodynamics area in our pursuit of finding more energy-efficient solutions for electromechanical systems or Hamiltonian systems in general.\\ 
In this effort, the efficiency of a two-link robot arm is considered.
The robot system is divided into two subsystems interacting with each other, and the cyclic averages of the energies in the system are considered. A classification problem is solved to restrict the trajectories of the robot arm in a way that the averaged energy transfer rate between the subsystems satisfies the thermodynamic principles. Afterward, a variational problem is defined to find the maximum fraction of the averaged energies of the subsystems necessary to satisfy the energy flow direction between them. 
\section{Energy Cyclo-Directionality}\label{ECDex}
In ~\cite{HaddadBook19}, two rules are introduced for interconnected systems that characterize the first and the second laws of thermodynamics.\\

Rule 1:
\begin{equation}
 \dot{E}_i=S_{i}-\sigma_{i}+\sum_{i,j\neq i} \phi_{ij}-S_{wi},\;\; \label{eq: ebal}
\end{equation}
where $E_{i} \geq 0$ are the energies of the subsystems, $\sigma_{i} \geq 0$ is the power dissipation to the environment, $\phi_{ij}$ is the net power received by subsystem $i$ from
$j\neq i$ and $S_{wi}$ and $S_{i}$ are the net rate of work performed
on the surrounding and the net external power supply received,
respectively. This rule is a representation of the first law of thermodynamics for interconnected systems. The second rule restricts
the direction of the energy flow in a thermodynamic system and is analogous to the second law of thermodynamics.\\

Rule 2:
\begin{equation}
\phi_{ij}(\beta_i E_i-\beta_j E_j) \leq 0 \label{eq: eflow}
\end{equation}
Where $\beta_i$ and $\beta_j$ are constants designed to act as a weight for the energies. 

By calculating the average quantities of any variable like $V$ using $\overline{V}_i=\frac{1}{T} \oint V_i(t) dt$ over the period T, the two mentioned rules can be re-written as follows:
\begin{equation}
 0=\overline{S_{i}}-\overline{\sigma_{i}}+\sum_{j \in \mathcal{I},j\neq i} \overline{\phi}_{ij}-\overline{S}_{wi},\;\;i,j \in \mathcal{I} \label{eq: ebalbar}
\end{equation}
\begin{equation}
\overline{\phi}_{ij}(\beta_i\overline{E}_i-\beta_j \overline{E}_j) \leq 0 \label{eq: eflowbar}
\end{equation}
Rules in this form are applicable to Hamiltonian systems and to the system at hand with two subsystems.\\

\section{System description}\label{ECDex}
\begin{figure} [H]
\begin{center}
\includegraphics[scale=0.28]{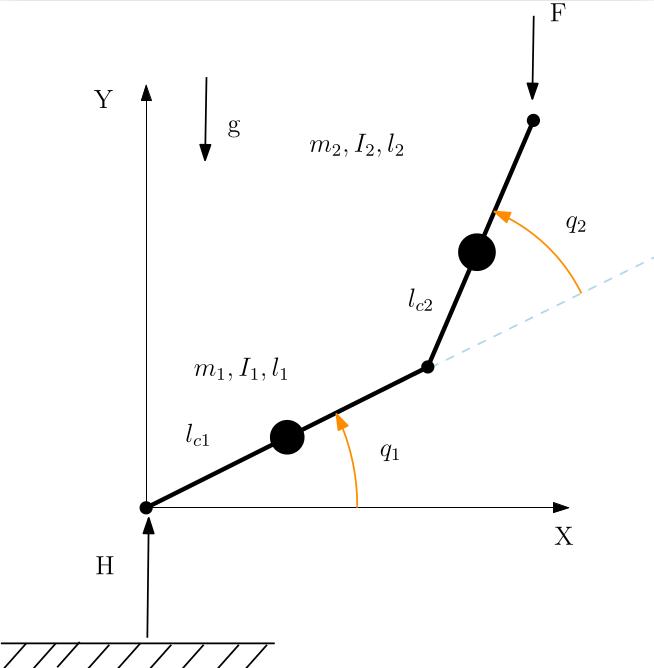} 

\caption{The two-link robotic arm} \label{fig: twolink}
\end{center}
\end{figure}

Figure \ref{fig: twolink} is a schematic of the two-link planner arm. The parameters are explained in Table \ref{tb: tbl1}:

\begin{table}[h!]

\centering
\renewcommand{\arraystretch}{1.5}
\begin{tabular}{ | m{2cm} | m{5cm}|} 
\hline
  $q_1$, $q_2$ & Angular positions of the joints  \\ 
  \hline
  $m_1$, $m_2$ & Masses of the links  \\ 
  \hline
  $F$ & External force (constant and vertical)  \\ 
  \hline
  $I_1$, $I_2$ & Moments of inertia of the links  \\ 
  \hline
  $l_1$, $l_2$ &  Lengths of the links  \\ 
  \hline
  $l_{c1}$, $l_{c2}$ & Lengths of the center of masses of the links  \\ 
  \hline
  $H$ & Reference level of the potential energy  \\ 
  \hline
\end{tabular} 
\caption{Physical parameters of the system}
\label{tb: tbl1}
\end{table}

Using Eq. (\ref{eq: ebal}) and writing the energy balance equation for link 1 we have:\\
\begin{equation}
\begin{split}
\dot E_1=\dot K_1+\dot V_1=\phi_{12}+\tau_1(t)\dot q_1(t)   \\
\dot E_2=\dot K_2+\dot V_2=\phi_{21}+\tau_2(t)\dot q_2(t)   \\
\end{split}
\label{eq: phi12cal}
\end{equation} 

Where $K_1$ and $V_1$ are the potential and kinetic energies of link 1 respectively and $\tau_i$s are the input torque at the joints. The kinetic energies of the links are in the form\\
\begin{equation}
\begin{split}
K_1=\dfrac{1}{2}\dot q^TD_1(q)\dot q,\:\:\: K_2=\dfrac{1}{2}\dot q^TD_2(q)\dot q \\
\end{split}
\label{eq: Ks}
\end{equation}
Where
\begin{equation}
\begin{split}
D_1(q)=
\begin{bmatrix}
D^{11}_1 & D^{12}_1 \\
D^{21}_1 & D^{22}_1 
\end{bmatrix} \\
D_2(q)=
\begin{bmatrix}
D^{11}_2 & D^{12}_2 \\
D^{21}_2 & D^{22}_2 
\end{bmatrix}
\end{split}
\label{eq: Ds0}
\end{equation}

\begin{equation}
\begin{split}
D^{11}_1=m_1l_{c1}^2 + I_1,\:\:D^{12}_1=0,\:\:D^{21}_1=0,\:\:D^{22}_1=0
\end{split}
\label{eq: Ds11}
\end{equation}

\begin{equation}
\begin{split}
D^{11}_2=m_1L_{c1}^2 + I_1\\
D^{12}_2=m_2 l_1^2 + 2m_2\cos(q_2) l_1 l_{c2} + m_2 l_{c2}^2 + I_2\\
D^{21}_2=m_2l_{c2}^2 + l_1 m_2 \cos(q_2) l_{c2} + I_2\\
D^{22}_2=m2l_{c2}^2 + i_2
\end{split}
\label{eq: ds}
\end{equation}
For the potential energies
\begin{equation}
\begin{split}
V_1=m_1 gH + l_{c1} m_1 g \sin(q_1)\\
V_2=m_2gl_1\sin(q_1) + l_{c2} \sin(q_1 + q_2) + m_2 gH
\end{split}
\label{eq: v1v2}
\end{equation}

 By assuming a periodic motion for the robotic arm and By integrating Eq. (\ref{eq: phi12cal}) over a cycle of motion, in the absence of dissipation and external power supply terms, and knowing that potential and kinetic energies would go back to their initial values at the end of the periodic motion, the averaged quantities would look like the following equation.
\begin{equation}
\begin{split}
0=\overline\phi_{12}+\oint \tau_1(t)\dot q_1(t) \,dt \rightarrow  \overline\phi_{12}= -\oint \tau_1(t)\dot q_1(t) \,dt\\
\end{split}
\label{eq: phi12calbar}
\end{equation} 
Which gives a formula for $\overline\phi_{12}$.
\subsection{The classification of the sign of $\phi_{12}$}
The task defined for the robotic arm is to move from point A to point B while a constant vertical force is acting at the end of the second link. 
To begin the optimization, first, it is necessary to restrict the space of trajectories to those that satisfy rule 2 in Eq.(\ref{eq: eflowbar}). To do that we need to know the sign of $\phi_{12}$ which is a function of the joint variables and system parameters $\mathcal{P}$:\\
\begin{equation}\label{eq: phifunc}
\phi_{12}=\phi_{12}(q,\dot q,\ddot q,\mathcal{P})\\
\end{equation} 
$q$, $\dot q$,  and $\ddot q$ are the joint position, velocity, and accelerations. \\
Finding an analytical solution to determine the sign of  $\phi_{12}$ is of a high complexity even for the simple systems at hand.  As an alternative approach, a Fourier series expansion of the joint angular positions $q_1$ and $q_2$ is considered (Eq. \ref{eq: q1} and Eq. \ref{eq: q2}), and a classification problem using a quadratic discernment model is solved to find the space of the  Fourier coefficients resulting in a specific sign of $\phi_{12}$. 
\begin{equation}\label{eq: q1}
q_1(t)=a_{10}+\sum_{j=1}^{m}a_{1j}cos(j\omega t)+\sum_{j=1}^{m}b_{1j}sin(j\omega t)\\
\end{equation} 
\begin{equation}\label{eq: q2}
q_2(t)=a_{20}+\sum_{j=1}^{m}a_{2j}cos(j\omega t)+\sum_{j=1}^{m}b_{2j}sin(j\omega t)\\
\end{equation} 
 
\begin{equation}\label{eq: phinew}
\omega=\dfrac{2\pi }{T}
\end{equation} 
$a_{ij}$ and $b_{ij}$ are the unknown coefficients and $m$ is the number of the terms considered for the Fourier series expansion. $T$ is the period of the motion and $\omega$ is the frequency. \\

The end effector is assumed to start moving from point $A$ and reach point $B$ at time $t=t_b< T$ and back to its original position in time $T$ while having zero velocity at those points. Solving for the joint variables at those points gives the joint velocities to be zero too. These constraints lead to the following equations:\\
\begin{equation}\label{eq: q0}
q_1(0)=q_1^a,\:\:\dot q_1(0)=0,\:\:q_2(0)=q_2^a ,\:\:\dot q_2(0)=0
\end{equation}
\begin{equation}\label{eq: qb0}
q_1(t_b)=q_1^b,\:\:\dot q_1(t_b)=0,\:\:q_2(t_b)=q_2^b ,\:\:\dot q_2(t_b)=0
\end{equation}
Where $(q_i^a,\dot q_i^a)$ and $(q_i^b,\dot q_i^b)$  are the angular position and angular velocities of the links at times $t=0,T$, and $t=t_b$ respectively.\\

For the classification problem, $m=4$ is chosen, so we have a total of $2\times(2m+1)=18$ Fourier coefficients to consider, but due to 8 constraints in equations (\ref{eq: q0}), and (\ref{eq: qb0}), only 10 of them are independent. The period, T is also chosen as a parameter, so the number of independent variables would end up being 11.\\
A training data set of the size of two million samples is produced using the MATLAB random function. Each sample consists of 11 items including the Fourier coefficients and the period $T$. The Quadratic decrement regression model is used for the problem. This model gives us a second-order polynomial as a function of all the existing features. There are two classes to be determined (positive and negative signs of $\phi_{12}$). The curve dividing the classes has the quadratic form:
\begin{equation}
\begin{split}
f(X) = L_0+L_{11}x_1+L_{12}x_2+...+L_{1n}x_n+\\
L_{21}x^2_1+L_{22}x^2_2+...+L_{2n}x^2_n+\\
L_{31}x_1x_2+L_{32}x_1x_3+...+L_{3n}x_{n-1}x_n
\end{split}
\label{eq: curve}
\end{equation}
Where $X$ is the vector containing the features and $L_{ij}$ are constant coefficients. $f(X)=0$ is the dividing line between classes.
The sign of $f(X)$ shows that on which side of the curve the samples lie. The sign of $\phi_{12}$ is equal to this sign. To make the predictions more accurate, the tolerance variable $\epsilon$ is defined to ignore the data points near the dividing line $f(X)=0$. The sign of $\phi_{12}$ would be determined by: \\
\begin{equation}\label{eq: phisign}
sign(\phi_{12}) =
    \begin{cases}
      1  & \text{$if\:f(X)>\epsilon\ $} \\
      -1 & \text{$if\:f(X)<-\epsilon $}\\
    \end{cases}
\end{equation}

\begin{figure} [H]
\includegraphics[width=\linewidth]{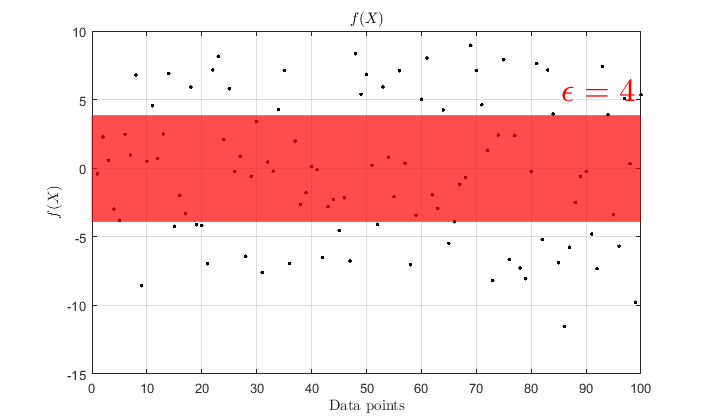} \caption{Points near line $f(X)=0$  are prone to give wrong answers. By defining a threshold variable like $\epsilon$, only the more reliable points farther from the boundary would be considered.} \label{fig: fxsign}
\end{figure}

Figure \ref{fig: fxsign} shows the value of the $f(X)$ for 100 data points and the tolerance band of $\epsilon=4$.

\subsection{Variational problem of the energy ratio}
By combining two constants $\beta_1$ and $\beta_2$ in Eq. (\ref{eq: eflowbar}) into one variable $\gamma$, the inequality can be rewritten in the following forms for a system with two subsystems.
\begin{equation}
\overline\phi_{12}(\gamma_{12} \overline E_1-\overline E_2) \leq 0.\:\:with\:\gamma_{12} = \dfrac{\beta_1}{\beta_2}\label{eq: eflow1}
\end{equation}
and
\begin{equation}
\overline\phi_{12}( \overline E_1-\gamma_{21} \overline E_2) \leq 0,\:\:with\:\gamma_{21} = \dfrac{\beta_2}{\beta_1}\label{eq: eflow2}
\end{equation}

To satisfy Equations (\ref{eq: eflow1}) and (\ref{eq: eflow2}), it would be necessary for $(\beta_1\overline{E}_1-\beta_2 \overline{E}_2)$ in Eq. (\ref{eq: eflowbar}) to have the correct sign based on the sign of $\overline\phi_{12}$. 
\begin{equation}\label{eq: signcom}
    \begin{cases}
       \text{$if\:\overline\phi_{12}<0\rightarrow (\gamma_{12}\overline{E}_1- \overline{E}_2)>0\rightarrow(\overline{E}_2-\gamma_{12} \overline{E}_1)<0 $} \\
       \text{$if\:\overline\phi_{12}>0\rightarrow (\overline{E}_1-\gamma_{21} \overline{E}_2)<0$} \\
    \end{cases}
\end{equation}

The $\gamma$ can be found using\\
\begin{equation}
\gamma_{ij}=\underset{T,q}{sup}\:\:\dfrac{\overline{E}_j}{\overline{E}_i} \label{eq: gamsuper}
\end{equation}
Where $q$ is the state vector and T is the period of the motion \cite{IEEELCS22}. $\gamma$ would be different in each case but as it is elaborated in the following sentences, for the two-link robotic arm, determination of a finite value for $\gamma$ is possible only for the second case where $\overline\phi_{12}>0$.\\

 In case $\overline\phi_{12}<0$ according to Eq. (\ref{eq: gamsuper}), $\gamma_{21}$ would be determined by setting $i=2$ and $j=1$. In other words, the minimum desired value for $\gamma_{21}$ would be determined by maximizing the averaged energy of link two divided by the averaged energy of link one. For this case, it is possible to imagine trajectories that would produce a fixed value for $\overline{E}_1$ while producing an infinite value for $\overline{E}_2$. For example by fixing the first link and giving a large angular velocity to the second link. That would make $\gamma_{12}$ go to infinity. So a bounded value for $\gamma_{12}$ is not possible.\\
 
 For the second case with $\overline\phi_{12}>0$, with $i=1$ and $j=2$ in Eq. (\ref{eq: gamsuper}), a finite value for $\gamma_{21}$ is reachable given that the value of $\overline{E}_1$ can not increase unboundedly independent of $\overline{E}_2$. An upper bound for the ratio $\dfrac{\overline{E}_1}{\overline{E}_2}$ can be calculated by maximizing the function $J$ below. 
\begin{equation}
J(q_1,\dot q_1,q_2,\dot q_2)= \dfrac{\overline{E}_1}{\overline{E}_2}=\dfrac{\dfrac{1}{T}\oint {E}_1 dt}{\dfrac{1}{T}\oint {E}_2 dt} \label{eq: var1}
\end{equation}
The problem described above is in the form of a problem solvable by calculus of variations. 
If the reference level for potential energy is selected low enough that the overall energies of the links are always positive, then the following inequalities hold.

\begin{equation}
J=\dfrac{\oint {E}_1 dt}{\oint {E}_2 dt} \leq \oint \dfrac{{E}_1}{{E}_2} dt \label{eq: e1e21}
\end{equation}
\begin{equation}
\dfrac{1}{J}=\dfrac{\oint {E}_2 dt}{\oint {E}_1 dt} \leq \oint \dfrac{{E}_2}{{E}_1} dt \label{eq: e1e22}
\end{equation}
According to the equations (\ref{eq: e1e21}) and (\ref{eq: e1e22}), maximizing J is equal to minimizing $\dfrac{1}{J}$ and that is equal to minimizing the right side of the Eq. (\ref{eq: e1e22}).
By solving the variational problem for $\dfrac{1}{J}$, we arrive at the following Eulers equation:
\begin{equation}
\begin{split}
\dfrac{d}{dt}(\dfrac{\partial E_2}{\partial \dot q})E_1^2-
\dfrac{\partial E_2}{\partial q}\dot E_1 E_1-
\dfrac{d}{dt}(\dfrac{\partial E_1}{\partial \dot q}) E_2 E_1-\\
\dfrac{\partial E_1}{\partial \dot q}(\dot E_1 E_1-2\dot E_1 E_2)-
\dfrac{\partial E_2}{\partial q}E_1^2+
\dfrac{\partial E_1}{\partial q}E_2 E_1=0\\
\end{split}
\label{eq: euler}
\end{equation}
Where\:\: $E_1=K_1+V_1$\:\:and\:\:$E_2=K_2+V_2$.\\
By substituting $K_i$s from Eq. (\ref{eq: Ks})  and solving the equation, it can be shown that $\dot q=0$ and $q_1=-\dfrac{\pi}{2}$  and $q_2=0$ satisfy Eq. (\ref{eq: euler}). It means this point is an extremum of the functional. To examine whether this point is a minimum or a maximum,  the calculation of the second variation of $\dfrac{1}{J}$ is required. To prove the extremum point to be a minimum, Jacobi sufficient condition must be met \cite{hounkonnou2011extremum}.  To use these conditions following form for the functional is assumed and the concept of conjugate points is used.\\

\begin{equation}
J=\int_{a}^{b}  F(t,q,\dot q) dt \label{eq: JF}
\end{equation}
And 
\begin{equation}
P=\dfrac{1}{2}F_{\dot q \dot q}\:\: and\:\: Q=(F_{\dot q \dot q}-\dfrac{d}{dt}F_{q \dot q})
\label{eq: PQ}
\end{equation}
Given that $q$ and $\dot q$ are two by one vectors:
\begin{equation}
F_{qq}=F_{q^i q^k},\:\: F_{q \dot q}=F_{q^i \dot q^k}\:\: and \:\:F_{\dot q \dot q}=F_{\dot q^i \dot q^k},\:\:i,k=1,2
\label{eq: FUU}
\end{equation}

\textbf{Jacobi's sufficient condition}: \textit{ The extremum $q$ corresponds to a minimum of the functional J,  if the matrix $P(t,q,\dot q)$ is positive definite along this extremum, and the open interval $[a\:\:b]$ contains no points conjugate to $a$.}\\

To conclude there are no points conjugated to $a$ in $[a\:\:b]$, the determinant below must not vanish in any points inside $[a\:\:b]$ other than at $t=a$.
\begin{equation}
\begin{vmatrix}
h_{11} & h_{12} \\
h_{21} & h_{22} 
\end{vmatrix}\neq 0
\label{eq: deth}
\end{equation}
Where $h_{ij}$s are the answers to the \textit{Jacobi system} of equations \cite{hounkonnou2011extremum} and satisfy the boundary conditions:
\begin{equation}
h_{ij}(a)=0,\:\:\dot h_{ii}=1,\:\:\dot h_{ik}(a)=0
\label{eq: hbc}
\end{equation}

\begin{equation}
-\dfrac{d}{dt}(P\dot h )+Qh=0
\label{eq: jacobisys}
\end{equation}
For the extremum point $q=\begin{bmatrix} -\dfrac{\pi}{2} \\ 0 \end{bmatrix}$, $P$ and $Q$ found to be:
\begin{equation}
P=\begin{bmatrix}
0.051 & 0.040 \\
0.040 & 0.033 
\end{bmatrix}\:\:
Q=\begin{bmatrix}
0.298 & 0.076 \\
0.076  & 0.076 
\end{bmatrix}
\label{eq: PQnum}
\end{equation}
The above values are calculated for the physical parameters below:
\begin{equation}
\begin{split}
m_1=1,\: m_2=1,\:
I_1=1,\: I_2=1,\:
L1=0.8,\:L_2=1,\: \\
L_{c1}=0.2 L_1 ,\:L_{c2}=0.25 l_2,\:
H=L_1+L_2
\end{split}
\label{eq: PQnum}
\end{equation}
By substituting the values of $P$ (which is positive definite) and Q into Eq. (\ref{eq: jacobisys}), solving for $h_{ij}$ and applying the boundary conditions for $t=a$ in Eq. (\ref{eq: hbc}), we have $h_{ij}=c_{ij}\:sinh(t)$ where $c_{ij}$ are constants. By substituting in \ref{eq: deth} the determinant would be:
\begin{equation}
\begin{vmatrix}
c_{11}sinh(t) & c_{12}sinh(t) \\
c_{21}sinh(t) & c_{22}sinh(t) 
\end{vmatrix}= 0\rightarrow sinh(t)=0\rightarrow t=0
\label{eq: hsol}
\end{equation}
This means that the solutions can only be zero at $t=0$ which in our system is equal to the lower band of integration ($a=0$). So there are no points conjugate to $t=0$ in the interval $[0\:\:T]$. For the static solutions at hand, $T=\infty$. In conclusion, the extremum point is a minimum.\\

\bibliographystyle{IEEEtran}
\bibliography{biblio}
\end{document}